\documentclass[letterpaper, 10 pt, journal, twoside]{ieeetran}
\IEEEoverridecommandlockouts                              

\usepackage{ifpdf}

\usepackage{xcolor}
\usepackage{color}
\usepackage{amsfonts} 
\usepackage{textcomp}
\usepackage{makecell}
\usepackage[]{hyperref}
\usepackage{cite}
\usepackage[font=small]{subcaption} 

   \usepackage[pdftex]{graphicx}
   \graphicspath{{figures/}{../jpeg/}}
  \DeclareGraphicsExtensions{.pdf,.jpeg,.png}
\usepackage{amsmath}
\usepackage{esvect}
\usepackage{mathtools}
\usepackage{multirow}
\usepackage{rotating}
\usepackage{siunitx}
\usepackage{array}
\usepackage[algo2e]{algorithm2e} 
\usepackage{algorithm}
\usepackage{algorithmic}
\usepackage{siunitx}
\usepackage{pifont}
\usepackage{booktabs}
\usepackage{colortbl} 

\usepackage{textcomp}

\usepackage{tabularray}

\begin{document}

\markboth{RAL. Preprint Version. May, 2025} {ZHANG \MakeLowercase{\textit{et al.}}: Manipulating Elasto-Plastic Objects With 3D Occupancy and Learning-Based Predictive Control} 
\title{\LARGE \bf
Manipulating Elasto-Plastic Objects With 3D Occupancy and Learning-Based Predictive Control
}

\author{Zhen Zhang$^{1,2}$, Xiangyu Chu$^{1,2}$, Yunxi Tang$^{1}$, Lulu Zhao$^{3}$, Jing Huang$^{1,2}$, \\Zhongliang Jiang$^{4}$, and K. W. Samuel Au$^{1,2}$ 
\thanks{Manuscript received: January 17, 2025; Revised: April 17, 2025; Accepted: May 20, 2025. This letter was recommended for publication by Editor Júlia Borràs Sol upon evaluation of the reviewers’ comments. (\textit{Corresponding author:  Xiangyu Chu)}}
\thanks{ 
This work was supported in part by the Multi-scale Medical Robotics Center, AIR@InnoHK and in part by Direct Grants (The Chinese University of Hong Kong) No. 4055245.
$^{1}$: Department of Mechanical and Automation Engineering, The Chinese University of Hong Kong, Hong Kong SAR, China.  $^{2}$: Multi-scale Medical Robotics Center, Hong Kong SAR, China. $^{3}$: School of Artificial Intelligence, Beijing Normal University, China. $^{4}$: Computer Aided Medical Procedures, Technical University of Munich, Germany.
}
} 

\maketitle

\begin{abstract}
Manipulating elasto-plastic objects remains a significant challenge due to severe self-occlusion, difficulties of representation, and complicated dynamics. This work proposes a novel framework for elasto-plastic object manipulation with a quasi-static assumption for motions, leveraging 3D occupancy to represent such objects, a learned dynamics model trained with 3D occupancy, and a learning-based predictive control algorithm to address these challenges effectively. We build a novel data collection platform to collect full spatial information and propose a pipeline for generating a 3D occupancy dataset. To infer the 3D occupancy during manipulation, an occupancy prediction network is trained with multiple RGB images supervised by the generated dataset. We design a deep neural network empowered by a 3D convolution neural network (CNN) and a graph neural network (GNN) to predict the complex deformation with the inferred 3D occupancy results. A learning-based predictive control algorithm is introduced to plan the robot’s actions, incorporating a novel shape-based action initialization module specifically designed to improve the planner’s efficiency. The proposed framework in this paper can successfully shape the elasto-plastic objects into a given goal shape and has been verified in various experiments both in simulation and the real world.

\begin{IEEEkeywords}
Deformable Object Manipulation, 3D Occupancy, Elasto-Plastic Objects
\end{IEEEkeywords}

\end{abstract}


\section{Introduction}
\IEEEPARstart{D}{eformable} object manipulation (DOM) is essential for many applications in the real world, such as household, industrial, and healthcare. These applications often involve handling various types of deformable objects, which can be broadly categorized into two major categories: thin-shell structural objects, such as clothes~\cite{wu2023learning, Mo_2023}, and ropes~\cite{tang}; and volumetric objects, such as plasticine~\cite{robocraft, diffsrl} and dough~\cite{qi2022learning}. In this paper, we focus on the robot manipulation of elasto-plastic volumetric objects, which deform elastically under small stress and undergo permanent plastic deformation under larger stress, and the objects' dynamics are assumed to follow a quasi-static motion model. Unlike rigid objects which can be represented by 6D pose~\cite{graspnet}, deformable objects pose challenges in compact state representation and dynamics modeling due to their high degrees of freedom, complex deformations, and significant self-occlusion. Although the geometric representation like surface-level point clouds can provide rich spatial information, it is sparse and semantics-lacking.
\begin{figure}[t]
    \centering
    \includegraphics[width=\linewidth]{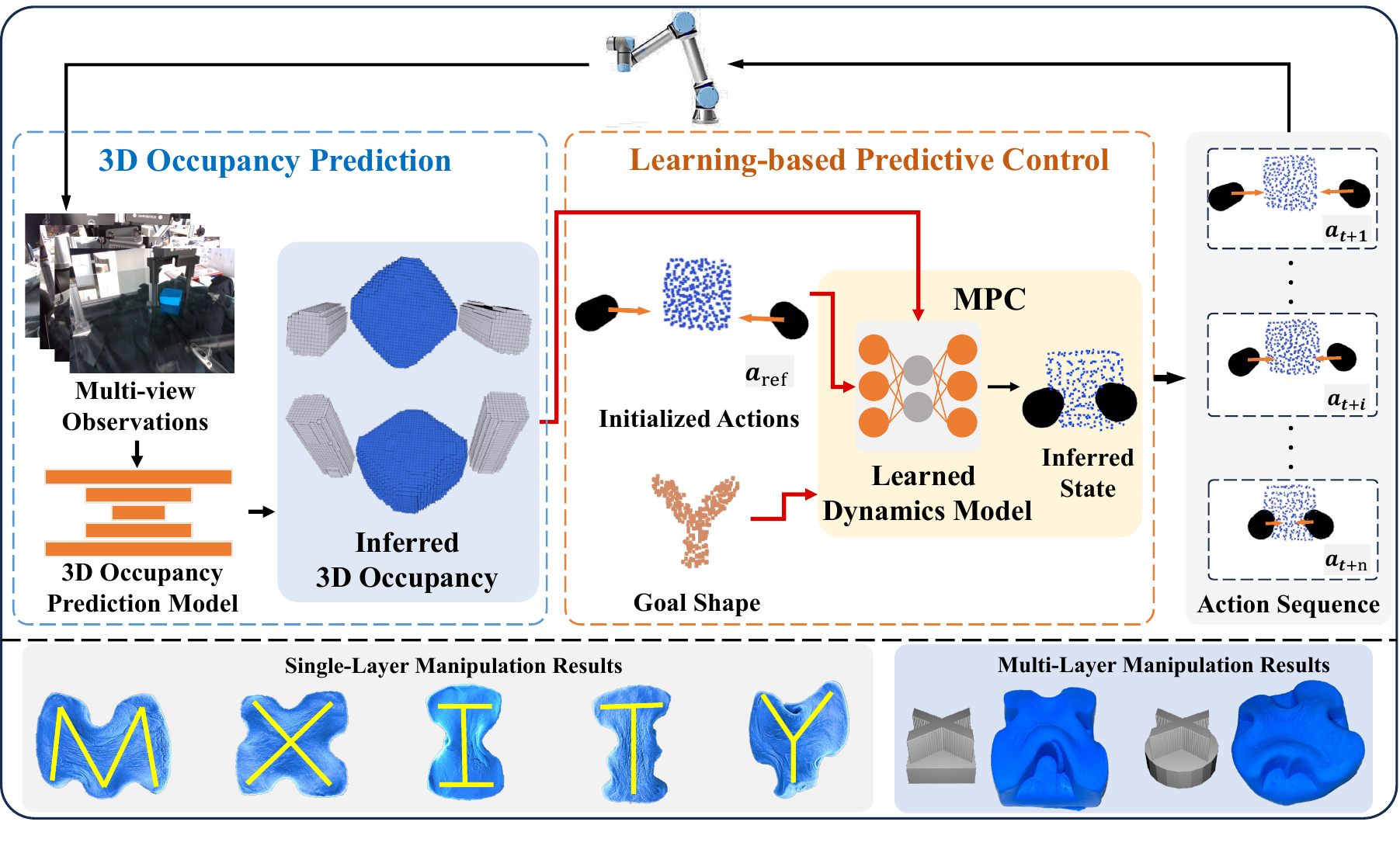}
    \caption{\textbf{Overview of our framework and results.}
    We predict 3D occupancy from RGB images, learn the dynamics with a deep neural network, and apply learning-based predictive control with shape-based action initialization to deform the object into a goal shape.
    }
    \label{whole_framework}
    \vspace{-1ex}
\end{figure}

In this work, we propose a novel framework to represent 3D deformable objects with dense voxel-grid 3D occupancy and develop modeling and control methods based on this representation. To demonstrate this paradigm, we choose 3D elasto-plastic objects (e.g., plasticine) as our study example, since their deformation is irreversible which poses a significant challenge in shaping them to a desired shape. Specifically, a novel elasto-plastic object manipulation framework is proposed, as shown in Fig.~\ref{whole_framework}. A 3D occupancy prediction network is proposed to generate inferred 3D occupancy from RGB images. Then, a learned dynamics model based on 3D occupancy is obtained via training a deep neural network empowered by 3D convolution neural network (CNN) and graph neural network (GNN). Finally, with a shape-based action initialization module, the model predictive control (MPC) is used to generate an action sequence for manipulation. Extensive experiments are conducted both in the simulator and the real world to evaluate our work. 
The key contributions are: 
\begin{itemize} 
\item We propose a novel framework to represent and manipulate elasto-plastic objects based on 3D occupancy, which includes a 3D occupancy prediction model, a learned dynamics model, and learning-based predictive control. 
\item We propose a shape-based action initialization module that uses the geometrical information between the current state and the goal to select actions for MPC to achieve the desired deformations. 
\item We build a novel and low-cost 3D data collection platform with a transparent operating plane to collect the data of 3D deformable objects with full spatial information, and propose a pipeline for 3D occupancy dataset generation of deformable objects from multi-view RGB-D. 
\end{itemize}
\section{Related Work}
\subsection{Deformable Object State Representation} 
In recent works, data-driven manipulation methods have drawn more attention to representing deformable objects in low dimensions. For some thin-shell deformable objects, such as cable and cloth, mesh embedding~\cite{tan2020realtime}, latent space features~\cite{matas2018sim}, and keypoint embedding~\cite{tang,deng2023learning} have been commonly used for their representation. For volumetric deformable objects like clay and tissue, particle-based and point cloud-based representations have gained popularity. For example, Li et al.\cite{li2018learning} and Shi et al.\cite{robocraft} proposed using particle-based representation to learn dynamics. Thach et al.\cite{thach2024defgoalnet, thach2023deformernet} and Bartsch et al.~\cite{sculptdiff} employed deep neural networks to process the point cloud of deformable objects. Although these representations can provide 3D spatial information, both particles and point clouds are sparse and semantics-lacking compared with RGB images. To bridge such a gap, Li et al.~\cite{li2024deformnet} introduced a representation model that combines a PointNet encoder with a conditional neural radiance field to process point clouds and RGB-Ds. However, this approach introduces the challenge of cross-modality feature fusion, i.e., efficiently integrating features from different modalities. In this work, we propose to use voxel-grid 3D occupancy, a predictable, fine-grained, semantic-rich, and inherently volumetric representation, to indicate the state of deformable objects and facilitate both dynamics modeling and planning for 3D DOM.

\subsection{Deformable Object Manipulation}
In addition to representation, modeling state transitions under action is critical for control. Prior works~\cite{suh2021surprising,xue2023neural,wang2023dynamic,zeng2021transporter,li2019learning} have explored diverse dynamics models for object manipulation, but face challenges when applied to complex 3D deformable objects. Methods such as Suh and Tedrake~\cite{suh2021surprising} and Transporter Networks~\cite{zeng2021transporter} relied on 2D image-space representations, limiting their ability to capture volumetric deformations and handle occlusions. Xue et al.\cite{xue2023neural} and Wang et al.\cite{wang2023dynamic} targeted granular materials, neglecting cohesive media like clay. The Koopman-based model~\cite{li2019learning} assumed globally linearizable dynamics, which is unrealistic for elasto-plastic behavior. Recent efforts have focused on learning graph-based dynamics models from low-dimensional features for 3D deformable object manipulation. Prior works~\cite{robocraft,robocook} utilized GNNs trained on down-sampled internal particles or surface mesh vertices, while Bartsch et al.\cite{sculptbot} proposed a GNN-based dynamics prediction pipeline using point cloud clusters. However, these methods either rely on accurate RGB-D-based mesh reconstruction during manipulation\cite{robocraft,robocook} or operate solely on surface-level information~\cite{sculptbot,robocook}, neglecting internal structural cues. This spatial information sparsity constrains contextual representation, ultimately degrading the performance of the learned dynamics. In contrast, our approach leverages dense occupancy-based representations and 3D CNNs to retain internal spatial information, enabling more accurate modeling of 3D deformable dynamics.

\section{Approach}
\subsection{Problem Formulation}
In this work, we aim to use a paralleled 2-finger gripper to manipulate an elasto-plastic object (i.e., plasticine) into a given goal shape ${s}_{goal}$ in a quasi-static setting with the observation from the multi-view cameras $I = \{I_{1}, I_{2},..., I_{n}\}$, where $n$ is the number of cameras. Specifically, we propose to use a dense, fine-grained, and volumetric representation inferred by a neural network with multi-view RGB images as input to indicate the state ${s}$ of plasticine, which can be represented as ${s} = \mathcal{N}(I)$, where $\mathcal{N}$ is a neural network. To manipulate the plasticine into the given goal shape ${s}_{goal}$, a sequence of actions $\mathcal{A} = \{a_{0}, a_{1}, ..., a_{t-1}\}$ will be applied upon the plasticine whose initial state is ${s_{0}}$, and the state of the plasticine will finally transit from ${s_{0}}$ to ${s_{t}}$, where ${s_{t}} \approx {s}_{goal}$ is expected and measured by quantitative metrics. The complex transition of plasticine can be learned with a neural network. Specially, given the inferred current plasticine state $s_{t-1}$ and gripper action $a_{t-1}$, the state $s_{t}$ at time step $t$ can be represented as:
\begin{equation}
    s_{t} = \mathcal{G}\left(s_{t-1}, a_{t-1}\right) = \mathcal{G}\left(\mathcal{N}\left(I_{t-1}\right), a_{t-1}\right)
\end{equation}
where $\mathcal{G}$ represents a trained neural network to approximate the state transition function. We can formulate the manipulation task as a model predictive control with the learned dynamics model. The cost function $\mathcal{L}$ is defined to measure the error between the current state $s_{t}$ and the given goal state ${s}_{goal}$. The sequence of actions $\mathcal{A} = \{a_{0},a_{1},...,a_{t-1}\}$ can be optimized by minimizing the cost function: $\mathcal{A}^{\ast}=\mathop{\arg\min}\mathcal{L}\left(\mathcal{G}\left(s_{0}, \mathcal{A}\right), s_{goal}\right)$.
\begin{figure}[t]
    \centering
    \includegraphics[width=0.8\linewidth]{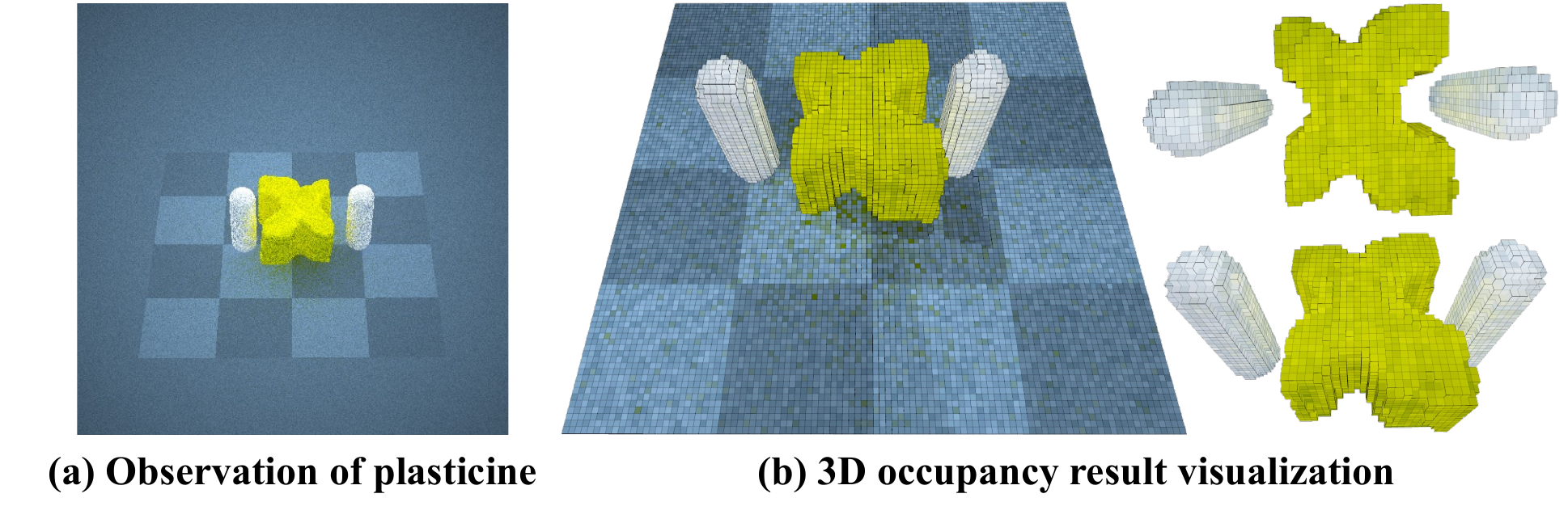}
    \caption{\textbf{Inferred 3D occupancy of manipulation scenario in the simulator.} (a) RGB observation. (b) Inferred 3D occupancy visualization. The grey, yellow, and dark blue parts represent the gripper, plasticine, and operating plane, respectively.}
    \label{sim_occ_sample}
\end{figure}
\subsection{3D Occupancy-based State Representation for DOs}
\begin{figure}[t]
    \centering
    \includegraphics[width=0.85\linewidth]{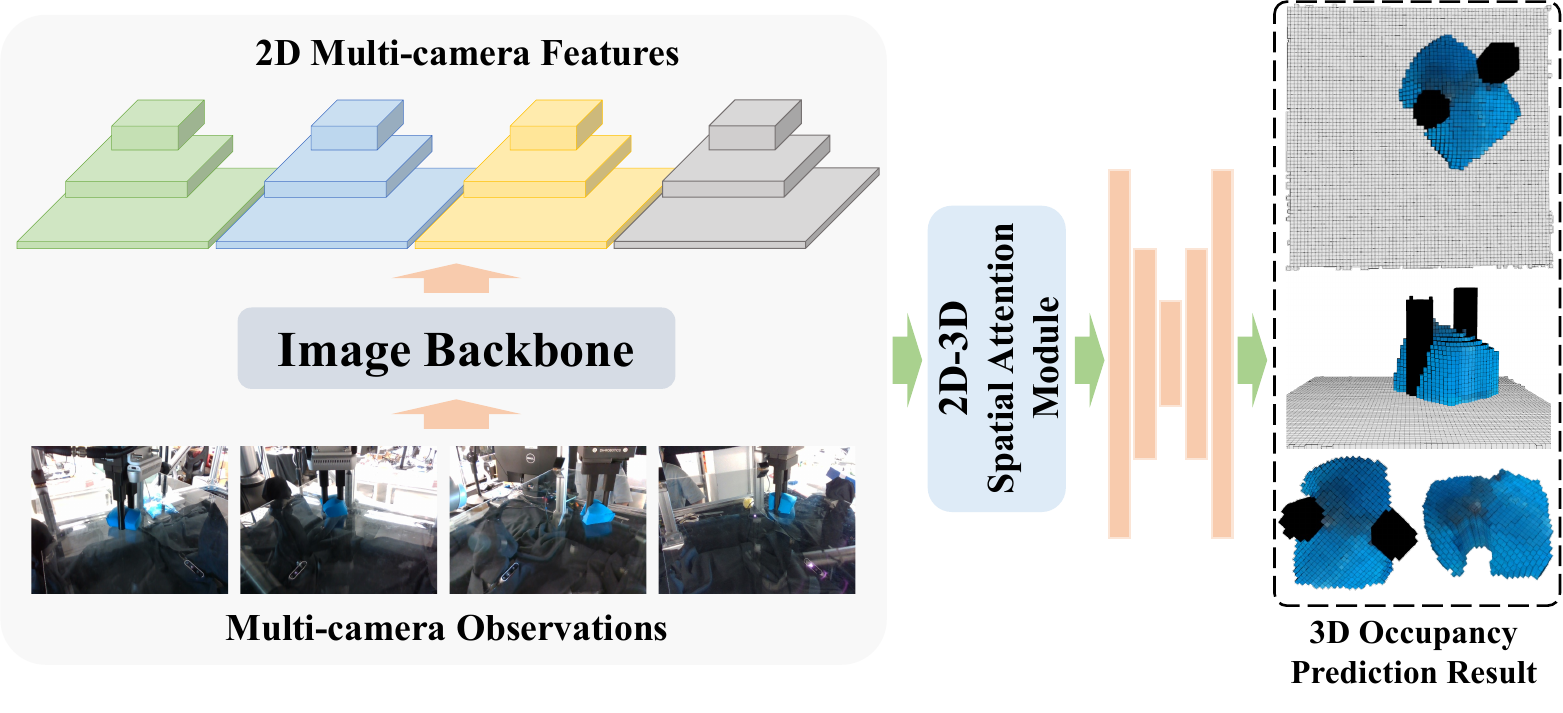}
    \caption{\textbf{3D occupancy prediction framework.} 
    We use four cameras to extract multi-scale features, fuse them with 2D-3D spatial attention, and predict 3D occupancy, supervised by the ground truth.
    }
    \label{occ_net}
\end{figure}

We utilize a voxel grid structure to represent the 3D occupancy of deformable objects. 3D occupancy is an effective representation for reconstructing multi-camera deformable object manipulation scenarios, as shown in Fig.~\ref{sim_occ_sample}. First, 3D occupancy is an inherently dense and fine-grained representation that can provide detailed, semantic (e.g., classes), and spatial information compared with RGB-D images and point clouds, which can be leveraged to extract high-dimensional features for learning the complex state transition dynamics model. Secondly, trained with occluded samples, the neural network can predict occluded areas between grippers and the deformable object according to the rich semantic information without providing depth information. Given multi-view RGB images $I^{t}$, where $t$ is the time stamp, the 3D occupancy prediction result can be represented as:
\begin{equation}
    \mathcal{O}_t =\mathcal{N}(I_{1}^{t}, I_{2}^{t},..., I_{n}^{t})
\end{equation}
where $\mathcal{O}_t$ is the 3D occupancy. Its value is between $0$ and $1$, representing the occupied probability of the voxel grids. Furthermore, we use the farthest point sampling (FPS) algorithm~\cite{fps} to down-sample the 3D occupancy $\mathcal{O}$ into a sparse voxel set $K=\left\{v_{0}, v_{1},..., v_{k}\right\}$ to indicate the 3D occupancy-based state $s$ of the plasticine. Therefore, the state of the plasticine can be represented as $s_{t}=FPS(\mathcal{O}_{t})=\left\{v_{0}^{t}, v_{1}^{t},..., v_{k}^{t}\right\}$.
The framework of 3D occupancy prediction is shown in Fig.~\ref{occ_net}. Specifically, we first use an image backbone network (i.e., ResNet-50~\cite{resnet}) to extract multi-scale features of multi-camera images $I$. The 2D-3D spatial attention module and multi-scale occupancy prediction proposed by~\cite{surroundocc} are used to fuse multi-camera features and output 3D occupancy prediction. The prediction network is supervised by the dense 3D occupancy ground truth generated by the pipeline proposed in Sec.~\ref{dom_pipeline_sec}. 
\begin{figure*}[tb]
    \centering
    \includegraphics[width=1.0\textwidth]{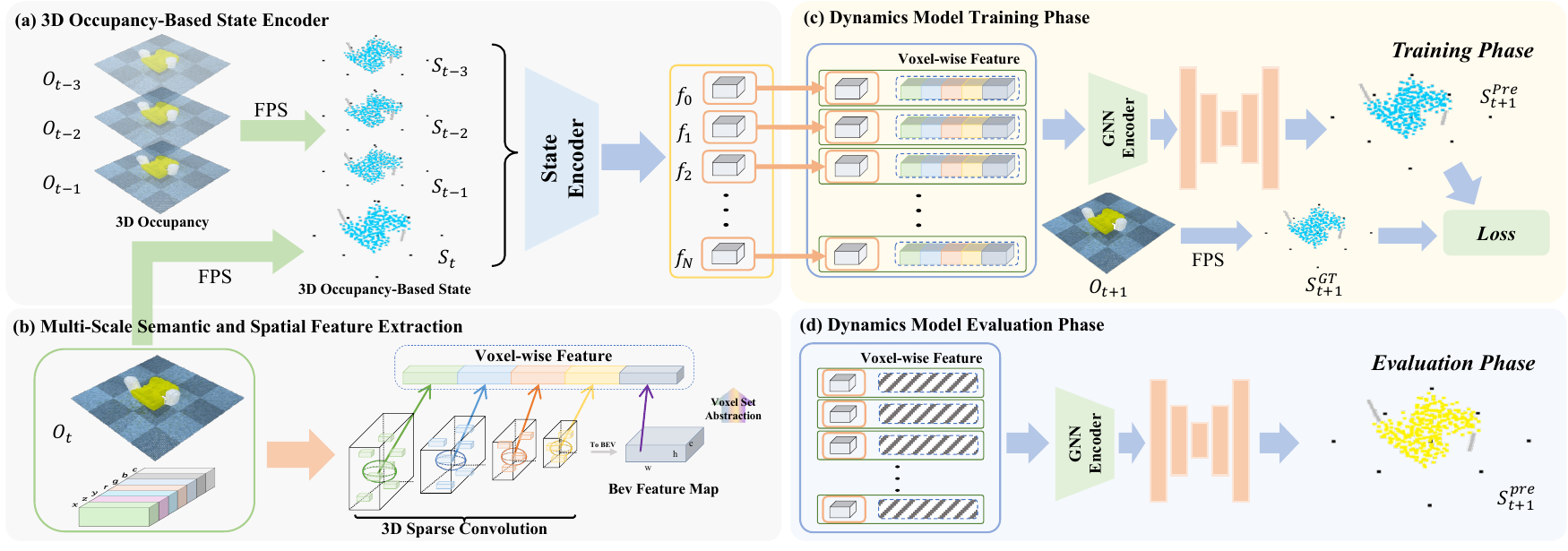}
    \caption{\textbf{Overall structure of our proposed 3D CNN-Based dynamics model.} (a) The 3D occupancy of the current step $t$ and previous three steps $t-3 \cdots t-1$ is down-sampled to construct a voxel-based state graph, and node features $f_{j}$ are extracted through a state encoder. (b) The 3D occupancy of the current step $t$ is fed into a 3D sparse CNN to learn multi-scale semantic and spatial features. The learned voxel-wise features of each graph node are then retrieved and summarized into a feature set from multiple levels through a voxel set abstraction module. (c) During the training phase, the aggregated node features are concatenated to the encoded node features $f_{j}$ and then fed into the GNN for dynamics model training. (d) During evaluation, voxel-wise features are masked and only used at the initial step ($t=0$).}
    \label{3D_conv}
\end{figure*}
\subsection{Learning-based Dynamics Model}
To model the plasticine's dynamics, we construct a voxel-based state graph by down-sampling from the 3D occupancy of the plasticine and use a deep neural network empowered by a 3D CNN and GNN, inspired by~\cite{robocraft}. To be specific, we first down-sample the 3D occupancy of the current state to construct a voxel-based state graph. Then, we use a 3D sparse CNN to extract the multi-scale features of the dense 3D occupancy, and use voxel set abstraction module~\cite{pvrcnn} to project the down-sampled voxels into the voxel grids and collect the features of each node for enhancing the spatial and semantic information. Lastly, we encode the node feature through a state encoder, and fuse the voxel-wised features and node features for the dynamics model learning. The overall structure is shown in Fig.~\ref{3D_conv}.
\subsubsection{Voxel-based State Graph Construction}
We construct a voxel-based state graph with the 3D occupancy-based state $s$, as $\mathcal{S}=(\mathcal{V},\mathcal{E})$. Its vertices $\mathcal{V}$ are the voxel set $K$ and the edges $\mathcal{E}$ between two vertices indicate their spatial relationship, which changes dynamically over time. For each vertex, all its neighbors within a predefined distance are connected as an edge, and each edge contains the receiver and sender particle index, object internal relation, and finger-to-object relation.
\subsubsection{Features Aggregation for Graph Nodes}
The spatial resolution of the voxel-grid 3D occupancy is $L\times W\times H$, where the features of the non-empty voxels are the 3D coordinates ${x,y,z}$, the colors ${r, g, b}$, and the class ${c}$. We use a plain sparse 3D CNN backbone network to extract the multi-scale semantic features. By default, the plain sparse CNN backbone network utilizes a series of $3\times3\times3$ 3D sparse convolution and has 4 levels, with the feature strides $\{1, 2, 4, 8\}$. We name the output sparse features $\{F_1, F_2, F_3, F_4\}$ of each level, respectively. For a voxel $v_{i} \in K$ with coordinate $(x_{v_{i}}, y_{v_{i}}, z_{v_{i}})$, the sparse features of the voxel from the last three levels $\{F_4, F_3, F_2\}$ can be represented as $f_4^{v_i} = F_4\left(\frac{x_{v_i}}{2^3}, \frac{y_{v_i}}{2^3}, \frac{z_{v_i}}{2^3}\right), f_3^{v_i} = F_3\left(\frac{x_{v_i}}{2^2}, \frac{y_{v_i}}{2^2}, \frac{z_{v_i}}{2^2}\right), f_2^{v_i} = F_2\left(\frac{x_{v_i}}{2^1}, \frac{y_{v_i}}{2^1}, \frac{z_{v_i}}{2^1}\right)$.
The 3D coordinate $(x_{v_{i}}, y_{v_{i}}, z_{v_{i}})$ of the voxel $v_{i}$ is projected to the 2D bird's-eye-view coordinate system, and we utilize bilinear interpolation to obtain the features $f^{bev}_{v_{i}}$ from the bird-view feature maps. Therefore, the multi-scale semantic feature $f_{v_{i}}$ for each voxel $v_{i}$ is generated by concatenating the features from different levels and BEV features following 
\begin{equation}
\begin{aligned}
    f_{v_{i}} = [f_4^{v_{i}}, f_3^{v_{i}}, f_2^{v_{i}}, f^{bev}_{v_{i}}], \text{for}\ i = 1, ..., n.
\end{aligned}
\end{equation}
\subsubsection{Loss Functions and Evaluation Metrics}
We use three loss functions to evaluate the similarity between the predicted and the ground truth state.

\textbf{Earth Mover’s Distance (EMD).} The Earth Mover’s Distance $L_{EMD}$ between two distributions $\mathcal{V}_{p},\mathcal{V}_{gt} \subseteq \mathbb{R}^3$ can be defined as $L_{EMD}(\mathcal{V}_{p},\mathcal{V}_{gt}) = \mathop{\min}\limits_{\mu:\mathcal{V}_{p}\rightarrow \mathcal{V}_{gt}}\sum_{x \in \mathcal{V}_{p}}\|x-\mu(x)\|_{2},$ where $\mu:\mathcal{V}_{p}\rightarrow \mathcal{V}_{gt}$ is the a bijection~\cite{emd}. Here, $\mathcal{V}_p$ and $\mathcal{V}_{gt}$ are assumed to have the same volume.

\textbf{Chamfer Distance (CD).}
Chamfer Distance $L_{CD}$ between two distributions $\mathcal{V}_{p},\mathcal{V}_{gt} \subseteq \mathbb{R}^3$ is: $L_{CD}(\mathcal{V}_{p},\mathcal{V}_{gt}) = \frac{1}{|\mathcal{V}_{p}|}\sum_{x \in \mathcal{V}_{p}}\mathop{\min}\limits_{y\in \mathcal{V}_{gt}}\|x-y\|_{2} + \frac{1}{|\mathcal{V}_{g t}|}\sum_{y \in \mathcal{V}_{gt}}\mathop{\min}\limits_{x\in \mathcal{V}_{p}}\|y-x\|_{2}.$

\textbf{Density-aware Chamfer Distance (DCD).} Since CD only considers its nearest neighbor in the other set while ignoring the surroundings,~\cite{wu2021density} proposed DCD. The Density-aware Chamfer Distance $L_{DCD}$ between two distributions $\mathcal{V}_{p},\mathcal{V}_{gt} \subseteq \mathbb{R}^3$ can be calculated: $L_{DCD}(\mathcal{V}_{p},\mathcal{V}_{gt}) = \frac{1}{2}(\frac{1}{\lvert\mathcal{V}_{p}\rvert}\sum_{x \in \mathcal{V}_{p}}(1-\frac{1}{n_{\hat{y}}^{\lambda}}e^{-\alpha||x-\hat{y}||_{2}}) + \frac{1}{\lvert\mathcal{V}_{gt}\rvert}\sum_{y \in \mathcal{V}_{gt}}(1-\frac{1}{n_{\hat{x}}^{\lambda}}e^{-\alpha||y-\hat{x}||_{2}}))$,
where $\hat{y}=\min_{y \in \mathcal{V}_{gt}}||x-y||_{2}$, $\hat{x}=\min_{x \in \mathcal{V}_{p}}||y-x||_{2}$, $n_{\hat{y}}^{\lambda}=|\mathcal{V}_{p}^{\hat{y}}|$, $n_{\hat{x}}^{\lambda}=|\mathcal{V}_{gt}^{\hat{x}}|$ and $\alpha$ denotes a temperature scalar.

The total loss in our experiments is the weighted sum of the three distance functions mentioned above: $L(\mathcal{V}_p, \mathcal{V}_{gt}) = w_1 \times L_{EMD}(\mathcal{V}_p, \mathcal{V}_{gt}) + w_2\times L_{DCD}(\mathcal{V}_p, \mathcal{V}_{gt}) + w_3\times L_{CD}(\mathcal{V}_p, \mathcal{V}_{gt})$.
\subsection{Learning-based Predictive Control}
After obtaining the learned dynamics model, the model predictive control is used to control the gripper to manipulate the plasticine.
\subsubsection{Action Space and Goal-Conditioned MPC}
We use a paralleled 2-finger gripper to manipulate the plasticine, and the action space of two fingers can be denoted by two fingers' geometry center positions as $\left\{p_r = (x_r,y_r,z_r), p_l = (x_l,y_l,z_l)\right\}$. The action space of the gripper can be defined as $\left\{x, y, z, r_z, l\right\}$. $r_z$ is the gripper's rotation about the $z$ axis (i.e., perpendicular to the operating plane), which is $\arctan(\frac{y_l-y_r}{x_l-x_r})$. There are no rotations about the $x$ and $y$ axes. We define $x = \frac{x_r + x_l}{2}$, $y = \frac{y_r + y_l}{2}$, $z = \frac{z_r + z_l}{2} =z_r=z_l$. $l$ represents the distance between 2 gripper fingers, which is $\|p_r - p_l\|_2$. 
 
Given a goal state $\mathcal{S}_{g}$ of the plasticine and the action sequence $\mathcal{A}_{0\rightarrow t-1} = \{a_{0}, a_{1},..., a_{t-1}\}$ sampled from the gripper action space, where $t$ is the time horizon, we denote the resulting state $\mathcal{S}$ changing after applying the control inputs (i.e., action sequence $\mathcal{A}_{0\rightarrow t-1}$) as $\mathcal{S}_{0 \rightarrow t}=\{\mathcal{S}_0, ..., \mathcal{S}_t\}$. The task is to obtain the actions that minimize the loss $\mathcal{L}(\mathcal{S}_t, \mathcal{S}_{g})$ between the actual state $\mathcal{S}_t$ and the target state $\mathcal{S}_{g}$. Thus, L-BFGS~\cite{lbfgs}, a gradient-based optimization algorithm, is used to optimize $\mathcal{A}_{0\rightarrow t-1}$ with the gradient of $\mathcal{L}(\mathcal{S}_t, \mathcal{S}_{g})$, which is the same as the training loss of our dynamics model.

\begin{figure}[t]
    \centering
    \includegraphics[width=0.9\linewidth]{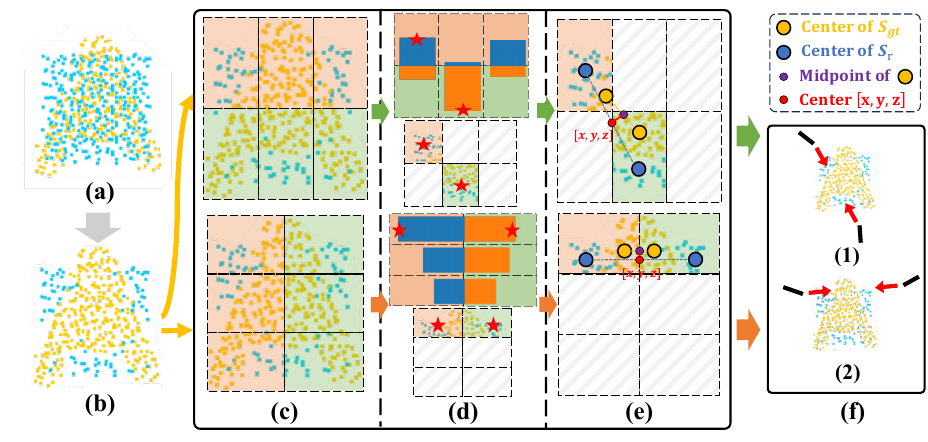}
    \caption{\textbf{Pipeline of our shape-based action initialization for the gripper.} The point cloud $s_{p}$ colored blue is the initial state of plasticine and the point cloud $s_{gt}$ colored yellow is the goal of plasticine (i.e., letter ``A''). (a) Align $s_{p}$ to $s_{gt}$. (a)$\rightarrow$(b) Filter out points close to $s_{gt}$ (c) Segment the point clouds into $N\times2$ regions along the $x$-axis and $y$-axis as two branches. (d) Calculate the cost of moving those points using Euclidean distance in each region, respectively. (e) Select two parts with the maximum cost from two branches respectively, and calculate the direction and center of the line connecting the two fingers as the reference initialization. (f) Visualization of initialized actions.}
    \label{shape_heur}
    \vspace{-1ex}
\end{figure}
\subsubsection{Shape-based Action Initialization}\label{action_initialization} While using MPC, we find that action initialization will significantly impact completion quality and repetition consistency as the deformation of plasticine is irreversible.
To this end, a shape-based action initialization module is proposed, as shown in Fig.~\ref{shape_heur}. First, we align the current state with the center of the target state (Fig.~\ref{shape_heur} (a)) and remove the point set $\{p_{0},p_{1}, \cdots, p_{i}\}\in s_{p}$ which is close to $s_{gt}$ (Fig.~\ref{shape_heur} (b)), to get the rest of point cloud $s_{r}$. Then, the point cloud is segmented into $m\times2$ regions along the $x$-axis and $y$-axis (Fig.~\ref{shape_heur} (c)) and calculate the cost of moving those points using Euclidean distance in each region respectively (Fig.~\ref{shape_heur} (d)), where $m$ is within the range of $3$ to $6$ in our experiments. Finally, we choose the two largest cost regions and obtain the center positions of point cloud $s_{r}$ and $s_{gt}$ in the two regions respectively. The rotation of the gripper $r_z$ is defined as the direction of the line connecting two center points of $s_{r}$, the center $x, y, z$ as the projection position of the midpoint of the line connecting two center points of $s_{gt}$ onto the line connecting the two center points of $s_{r}$ (Fig.~\ref{shape_heur} (e)). Finally, we can get the initial actions for MPC (Fig.~\ref{shape_heur} (f)).
\section{3D Occupancy Ground Truth For DOM}\label{dom_pipeline_sec}
In our experimentwe find that the network supervised
by RGB-D images is unable to predict dense enough occupancy. Thus, we need to generate dense occupancy labels for training. To this end, we built a novel data collection platform and proposed a pipeline to generate fine-grained dense occupancy ground truth based on raw RGB-D images.
\subsection{Data Collection Platform}
\begin{figure}[t]
    \centering
    \includegraphics[width=0.8\linewidth]{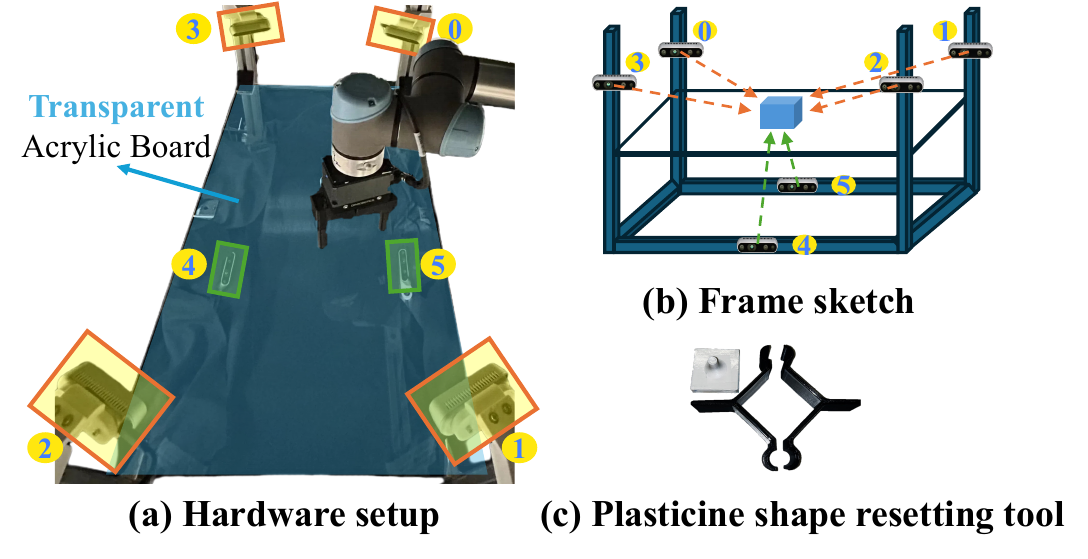}
    \caption{\textbf{Hardware setup of our data collection platform.} The blue part is the acrylic board serving as the operating plane. The orange parts are four cameras above the plane for collecting the top and side images. The green parts are two cameras below the plane for collecting the bottom images.}
    \label{platform}
    \vspace{-1ex}
\end{figure}

Conventional data collection platforms~\cite{pokeflex} can only collect the data of the objects above the plane while ignoring the bottom information. In practice, RGB-D images from the above views are unable to provide enough point clouds to build an enclosed mesh.
To address it, we used aluminum profiles to build a frame and installed a transparent acrylic board in the middle of the frame as the operating plane, allowing us to collect full spatial information (i.e., top, side, and bottom information) of the manipulated object. Six cameras are installed at different positions and angles to refine the mesh reconstruction. Specifically, we set 4 cameras on top of it and 2 cameras on the bottom to collect multi-view calibrated images of the manipulation scenario to obtain dense point clouds, as shown in Fig.~\ref{platform}. The bottom-view observations significantly enhance mesh reconstruction and occupancy quality by capturing occluded regions that top and side views miss. To minimize distortion, we use a high-transmittance acrylic board and depth sensors with robust filtering and careful calibration. For more details, we refer readers to~\cite{zhang2024dofs}.

\subsection{Data Collection and Processing}
In real-world scenarios, we collect multi-view RGB-D images and then convert them into dense point clouds $P=\{ P_{1}, P_{2}, \cdots, P_{n}\}$ with RGBs using the extrinsic and intrinsic parameters, where $n = 6$. We use the physical prior (i.e., size) of the platform to remove the points far from the platform and restrain the point cloud in the range of the platform (i.e., the platform size $[l, w, h]$ is $[\SI{1.0}{\metre}, \SI{1.0}{\metre}, \SI{0.2}{\metre}]$). Therefore, the raw point cloud of the manipulation scenario can be updated with $P_{raw}=M(P)$, as shown in Fig.~\ref{pipeline} (a), where $M$ is the mask to filter the points out of the platform. For the manipulation task, the region of interest (ROI) is only related to the deformable object and the gripper, so we cropped the point cloud $P_{raw}$ to $P_{roi}$, as shown in Fig.~\ref{pipeline} (b). We use color information as a color-based filter to remove the noisy points and segment the point cloud $P_{do}$ of the plasticine and the point cloud $P_{g}$ of the gripper from $P_{roi}$. For $P_{g}$, DBSCAN~\cite{ester1996density} is used to cluster the point clouds belonging to two primitives $\{P^{1}_{g}, P_{g}^{2}\}$ of the gripper. 

\subsection{Refinement with Geometry Prior}
Although a color-based filter is implemented, some missed points $P_{m}$ belonging to two primitives remain in $P_{do}$ due to occlusion and reflection. This is worse when the primitives are in contact with the plasticine. So we use $P_{xyz}$ and $[R, L]$ of two primitives to refine the point cloud of the plasticine, where $P_{xyz}$ is the 3D position of two primitives in the cartesian coordinate system in 3D space, $R$ and $L$ is the radius and the length of the primitive, respectively. We implement tiny dilation to remove the missed points of the two primitives $[P^{1}_{m}, P^{2}_{m}]$ in $P_{do}$. Therefore, the representation of the two primitives and the DO will be updated as $\{P^{1}_{g} = [P^{1}_{g}, P^{1}_{m}], P_{g}^{2}= [P_{g}^{2}, P^{2}_{m}]\}$ and $P_{do}=\{P_{do} - [P^{1}_{m}, P^{2}_{m}]\}$, respectively.
\subsection{Surface Reconstruction}
\begin{figure}[t]
    \centering
    \includegraphics[width=0.8\linewidth]{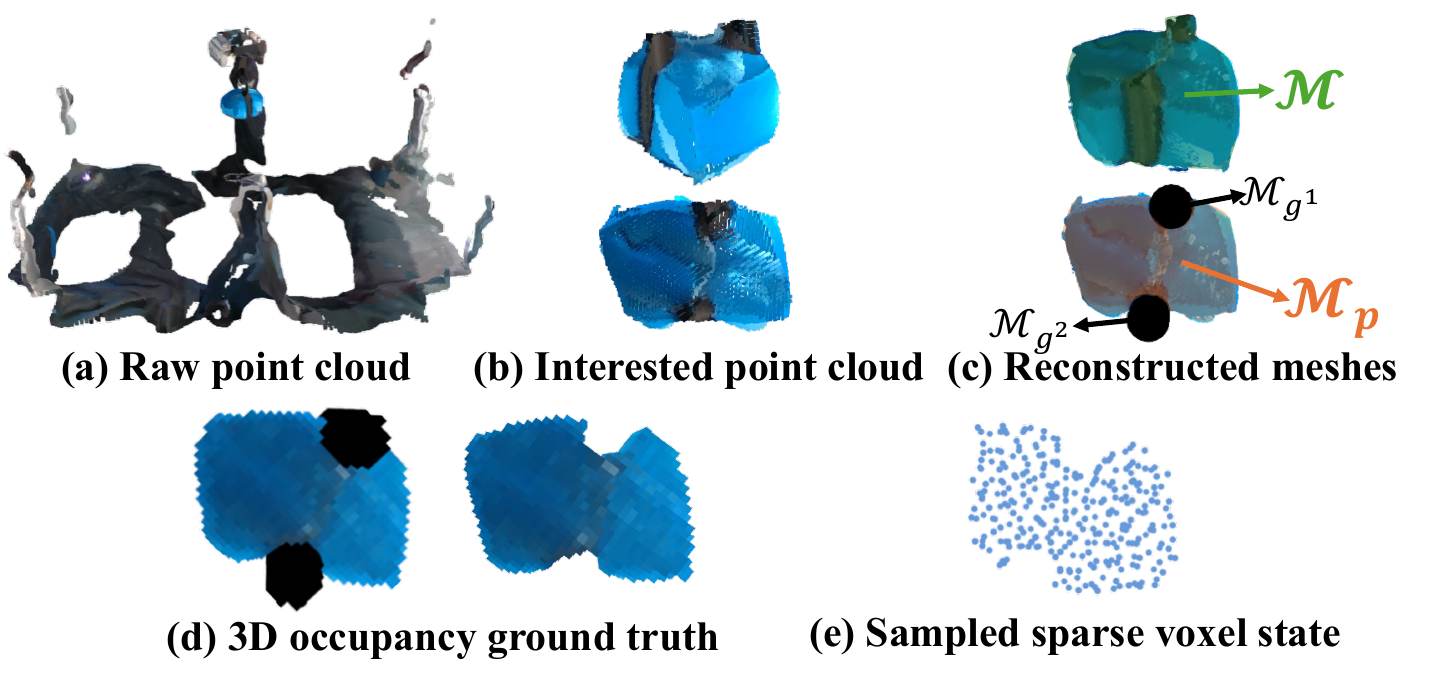}
    \caption{\textbf{Pipeline for occupancy ground truth generation.} (a) Converting RGB-Ds into dense and well-registered point clouds $P_{ra w}$ filtered by $M$. (b) Point cloud $P_{roi}$ of DO and gripper. (c) Reconstruct deformed meshes $\mathcal{M}$, $\mathcal{M}_{p}$, $\mathcal{M}_{g^{1}}$ $\mathcal{M}_{g^{2}}$. (d) Dense 3D occupancy ground truth. (e) Down-sampled voxel state from 3D occupancy.}
    \label{pipeline}
    \vspace{-1ex}
\end{figure}

Instead of only considering the gripper and plasticine as one point cloud to construct a whole mesh $\mathcal{M}$, we also construct the meshes of plasticine $\mathcal{M}_{p}$ and two primitives $\{\mathcal{M}_{g^{1}}$, $\mathcal{M}_{g^{2}}\}$ using their point cloud $P_{do}$, $P^{1}_{g}$ and $P^{2}_{g}$ shown in Fig.~\ref{pipeline} (c), via Poisson Surface Reconstruction~\cite{psr}, a learning-free method that uses smoothness prior of surface geometry to create watertight surfaces from oriented point sets. 
\subsection{Dense Occupancy Generation and Semantic Labeling}
The obtained mesh $\mathcal{M = \{V, E\}}$ fills up the holes of point clouds with evenly distributed vertices $\mathcal{V}$, so we can further convert the mesh into dense voxel set $V_{d}$ shown in Fig.~\ref{pipeline} (d). Specifically, first, the range $\{[x_{min},y_{min},z_{min}],[x_{max},y_{max},z_{max}]\}$ of  $\mathcal{M}$ is obtained from the evenly distributed vertices $\mathcal{V}$. Second, the space $\mathcal{R}=\{x_{min} \leq x \le x_{max}, y_{min} \leq y \le y_{max}, z_{min} \leq z \le z_{max}\}$ is voxelized with a fixed voxel size $[s_x, s_y, s_z]$.
After voxelization, we compute the signed distance function (SDF) of $\mathcal{M}$ for each voxel and remove the voxels outside $\mathcal{M}$. Finally, we use the plasticine mesh $\mathcal{M}_{p}$ and primitives $\{\mathcal{M}_{g^{1}}$ $\mathcal{M}_{g^{2}}\}$ to label the voxels inside them repeating the previous step. For the operating plane, we consider it as a thin shell, a cube of size $[l, w, h]$, where ${l}$, ${w}$, ${h}$ is the length, width, and height, and  ${h}$ equals to the voxel size $s_z$. The color of each voxel is defined as the average of the RGB values of the six vertices obtained by projecting the voxel center onto the mesh surface along the $x$, $y$, and $z$ axis. Following the proposed method, we can get dense volumetric occupancy with fine semantic labels without huge human efforts of annotation.

\section{Experiments and Results}
In this section, we present a comprehensive set of experiments to evaluate the effectiveness of our proposed method. Specifically, we evaluate the quality of state representation derived from voxels down-sampled from the predicted 3D occupancy and the performance of the dynamics model trained on the representation. Additionally, we perform ablation studies to highlight the benefits of fusing dense 3D occupancy features and the proposed shape-based action initialization module. Finally, we validate the performance of our method through extensive hardware experiments, showcasing its ability to manipulate plasticine into a given target shape.
\subsection{Experimental Setup}
\subsubsection{Simulation Setup}
The simulation~\cite{plasticinelab} features two capsule-shaped fingers, each with a radius of \SI{0.045}{\meter} and a length of~\SI{0.25}{\meter}, interacting with a square plasticine of size~$0.25\times0.25\times0.2$~\si{\meter}. Six RGB-D cameras are strategically placed at diverse viewpoints, including top and bottom perspectives, to capture RGB-Ds at a resolution of $512\times512$.
\subsubsection{Hardware Setup}
As shown in Fig.~\ref{platform} (a), we build this platform to conduct real-world experiments. We utilize a 6-DoF UR5e robotic arm equipped with a gripper (DH-Robotics PGI-140-80), and six RGB-D cameras (Real-sense D453i). Specifically, we 3D-printed two fingers for the gripper, each with a length of \SI{0.15}{\metre} and a radius of \SI{0.0075}{\metre}. Additionally, we calibrated the six RGB-D cameras, which operate at a frame rate of \SI{6}{\hertz} and a resolution of $640 \times 480$. For better understanding, a schematic of the platform is provided in Fig.\ref{platform} (b). Additionally, we designed and 3D-printed a specialized tool, illustrated in Fig.\ref{platform} (c). We control the UR5e robotic arm and the gripper under the Robot Operating System (ROS). The transformations between the two fingers and the robot base are published at a frequency of \SI{15}{\hertz}.
\subsubsection{Dataset Collection and 3D Occupancy Ground Truth Generation}
In the simulator~\cite{plasticinelab},we collected a total of $70$ episodes, comprising $8400$ frames, where $50$ episodes are used for training and $20$ episodes are for evaluation. In the real-world experiments, we collect $50$ episodes, consisting of $6000$ frames, as training data to fine-tune the pre-trained model for the real-world experiments. Specifically, each frame includes six RGB-D images and the positions of the two fingers w.r.t. the robot arm’s base frame. During each episode, the plasticine was pinched three times, with data recorded when the gripper and plasticine were aligned on the same plane. To ensure consistent initial conditions, a plasticine shape resetting tool shown in Fig.~\ref{platform} (c) is used to reshape the plasticine to a uniform starting shape and localized at the same position by clamping the tool to the gripper before each episode. Following the pipeline described in Sec.~\ref{dom_pipeline_sec}, we generated the 3D occupancy ground truth using six RGB-Ds to supervise the training of the 3D occupancy prediction model. 
\subsection{Experimental Results}
\subsubsection{3D Occupancy-based State Representation}
To model the plasticine dynamics, we down-sample the 3D occupancy of the plasticine to get a sparse voxel state as shown in Fig.~\ref{pipeline} (e) and construct a voxel-based state graph. First, we train a 3D occupancy prediction network using our generated 3D occupancy dataset on an Nvidia GeForce RTX-4080 with 16G memory. The occupancy prediction range is $[\SI{-0.1}{\metre}, \SI{0.1}{\metre}]$ for $X, Y$ axis and $[\SI{-0.01}{\metre}, \SI{0.07}{\metre}]$ for $Z$ axis. The shape of predicted occupancy is $100\times100\times40$ with the voxel size of $[\SI{0.002}{\metre}, \SI{0.002}{\metre}, \SI{0.002}{\metre}]$. Only specific parts of predicted occupancy are labeled as the plasticine, operating plane, and gripper, while the others would be considered noise. Then the trained model is employed to infer the 3D occupancy from four camera views, which is downsampled into a voxel set $K$ ($k=300$) representing the 3D occupancy-based state of the plasticine. The similarity between the down-sampled results and the ground truth indicates the quality of state representation, which will determine the performance of the trained dynamics model. So we compare the 3D occupancy-based method with two baseline methods proposed in~\cite{robocraft}, patch-based and crop-based method (four camera views) on the simulation dataset, averaged over $160$ frames (i.e., four pinches) using EMD, DCD, and CD. Specifically, the patch-based method patches the sampled results using points from the previous frame, and the crop-based method directly crops out the object point cloud and then down-samples it. For the 3D occupancy-based method, ``Prediction'' refers to the state that is down-sampled from the inferred occupancy with four cameras, while ``Ground Truth'' denotes the state that is down-sampled from the generated occupancy dataset with six cameras. Table.~\ref{Samplequality} shows that the 3D occupancy-based method has a more precise representation, achieving lower loss across all metrics.
\begin{table}[t]
    \centering
    \renewcommand\arraystretch{1.0}
     \scalebox{0.8}
    {
    \begin{tabular}{*{5}c}
        \hline
         \multicolumn{2}{c}{\multirow{2}{*}{\textbf{Methods}}} & \multicolumn{3}{c}{Average Over Four Pinches} \\ & & $\mathbf{EMD}_{\times1e^{-3}}$~$\downarrow$ & $\mathbf{DCD}_{\times1e^{-4}}$~$\downarrow$ & $\mathbf{CD}_{\times1e^{-3}}$~$\downarrow$ \\
        \hline
         \multicolumn{2}{c}{Patch-Based} & $32.7\pm3.86$ & $39.4\pm2.05$ & $38.6\pm2.12$\\ 
         \multicolumn{2}{c}{Crop-Based} & $30.5\pm1.62$ & $36.7\pm1.26$ & $37.1\pm0.953$ \\
         \hline 
         \multirow{2}{*}{\textbf{3D Occ-based}} & Prediction & ${28.6 \pm 1.23}$ & ${34.5 \pm 1.03}$ & ${36.0\pm 0.667}$ \\
         \cline{2-2}
         & \textbf{Ground Truth} & $\mathbf{27.9\pm1.14}$ & $\mathbf{34.2\pm0.998}$ & $\mathbf{35.8\pm0.676}$ \\ 
         \hline
    \end{tabular}
    }
    \captionsetup{font=small}
    \caption{\textbf{Quality comparison of different state representation methods.} We compared patch-based, crop-based, and 3D occupancy-based (3D Occ-based) methods in terms of EMD, DCD, and CD, respectively, over four pinches (160 frames).}
    \label{Samplequality}
\end{table}
\subsubsection{3D CNN-GNN Dynamics Model Training}
Based on our experiments, we set the hyper-parameters of $L_{DCD}$ as $\alpha=500$ and $\lambda=0.5$ for evaluation, and $\lambda=0.1$ and $\alpha=20$ for training. Additionally, the hyper-parameters of total loss $L(\mathcal{V}_p, \mathcal{V}_{gt})$ are selected as $w_1=0.5$, $w_2=0.4$ and $w_3=0.1$. We train our model with 3D occupancy-based state representation for 24 epochs and evaluate the performance of the learned dynamics model.
We compare our method against two strong baselines introduced in RoboCraft~\cite{robocraft} and RoboCook~\cite{robocook}. Both baselines employ GNN-only architectures and are trained for 100 epochs only using EMD and CD. The state representation in RoboCook is referred to as the surface-based method, where the representation is obtained by down-sampling the vertices of the object’s surface mesh. All the dynamics models used in our comparisons can be summarized as follows:
\begin{itemize}
\item RoboCraft \cite{robocraft}: Crop-based (state) + GNN (model) 
\item RoboCook \cite{robocook}: Surface-based (state)  + GNN (model) 
\item Ours: 3D Occ-based (state)  + 3D CNN-GNN (model) 
\end{itemize}
The evaluation results using EMD, DCD, and CD are shown in Table.~\ref{dynamics_result}, and our model trained with a 3D occupancy-based state representation consistently outperforms baselines across all metrics. These results can be attributed to the structural advantages of our representation. RoboCook uses surface-only nodes with limited connectivity, while our 3D occupancy includes internal nodes, enabling richer structural understanding. RoboCraft considers internal structure, but its crop-based method yields a sparse, local receptive field that lacks global spatial context. In contrast, our dense 3D features capture multi-scale spatial context, leading to more accurate predictions of the dynamics during manipulation. Note that, for the evaluation of the surfaced-based method in~\cite{robocook}, the mesh is reconstructed from surface-level points, and $k=300$ internal points are then down-sampled for the quantitative comparison.  

\begin{table}[tb]
    \centering
    \renewcommand\arraystretch{1.0}
    \scalebox{1.0}{
    \begin{tabular}{cccccc}
        \hline
        \multicolumn{2}{c}{\multirow{2}{*}{\textbf{Methods}}} & \multicolumn{3}{c}{Dynamics Model Average Evaluation Loss} \\ & & $\mathbf{EMD}_{\times1e^{-3}}$~$\downarrow$ & $\mathbf{DCD}_{\times1e^{-4}}$~$\downarrow$ & $\mathbf{CD}_{\times1e^{-3}}$~$\downarrow$ \\
        \hline
        \multicolumn{2}{c}{RoboCraft} & $24.3\pm2.82$ & 27.3$\pm$2.30 & 34.7$\pm$1.64 \\
        \multicolumn{2}{c}{RoboCook} & $25.5\pm2.92$ & $33.2\pm 2.02$ & $36.3\pm1.45$ \\
        \multicolumn{2}{c}{\textbf{Ours}}& $\mathbf{22.8\pm2.30}$ & $\mathbf{25.3\pm1.52}$ & $\mathbf{33.5\pm0.951}$ \\
        \hline
    \end{tabular}
    }
    \captionsetup{font=small} 
    \caption{\textbf{Mean and standard deviation of the dynamics model's performance comparison.} 
    }
    \label{dynamics_result}
\end{table}
\begin{figure}[tb]
    \centering
    \includegraphics[width=0.85\linewidth]{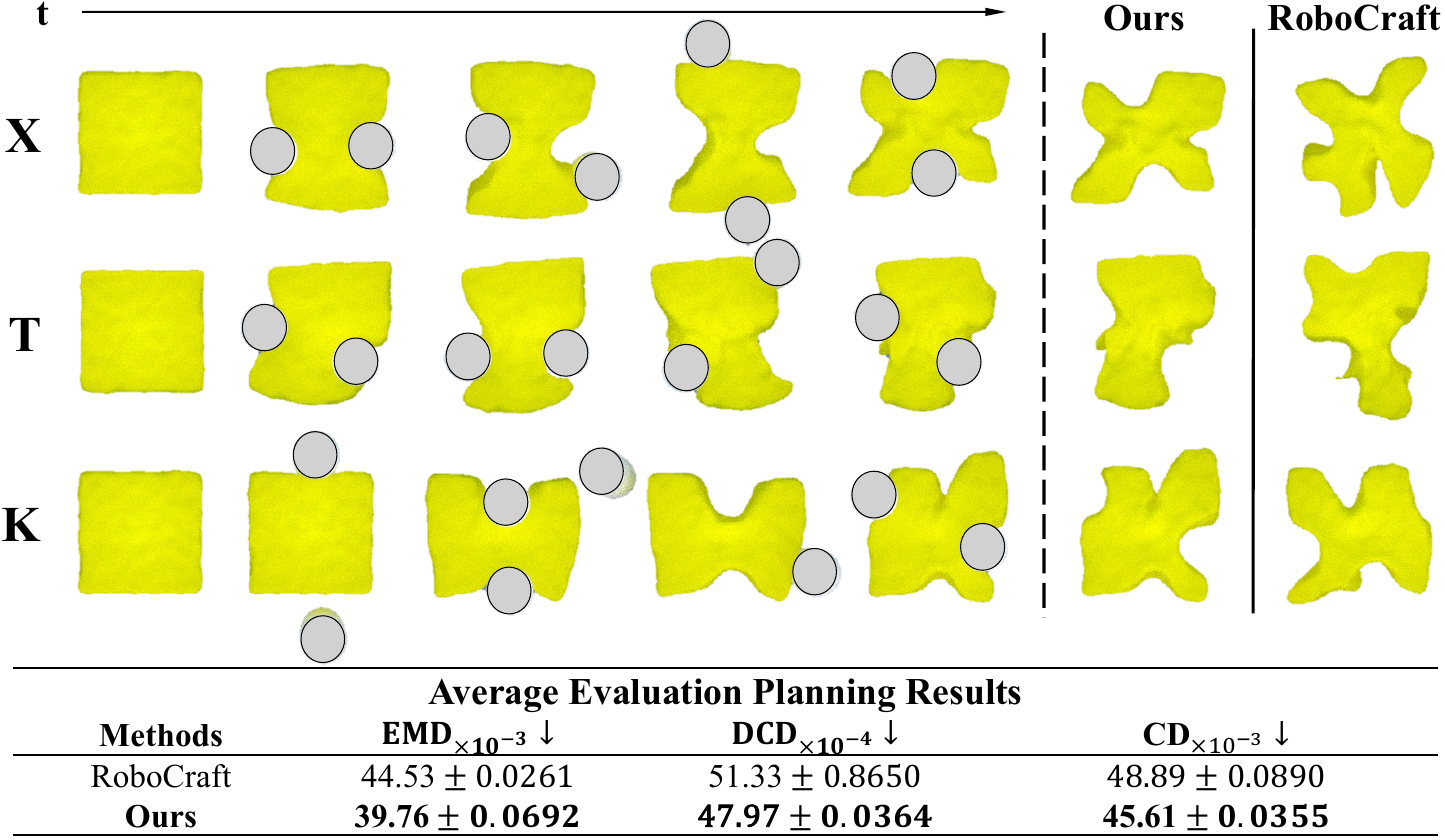}
    \caption{
    \textbf{Qualitative and quantitative results from three simulation trials.} Trials are performed with letters `X', `T', and `K'.}
    \label{sim_result}
\end{figure}
\subsubsection{Manipulation Results}
\begin{figure}[tb]
    \centering
    \includegraphics[width=0.9\linewidth]{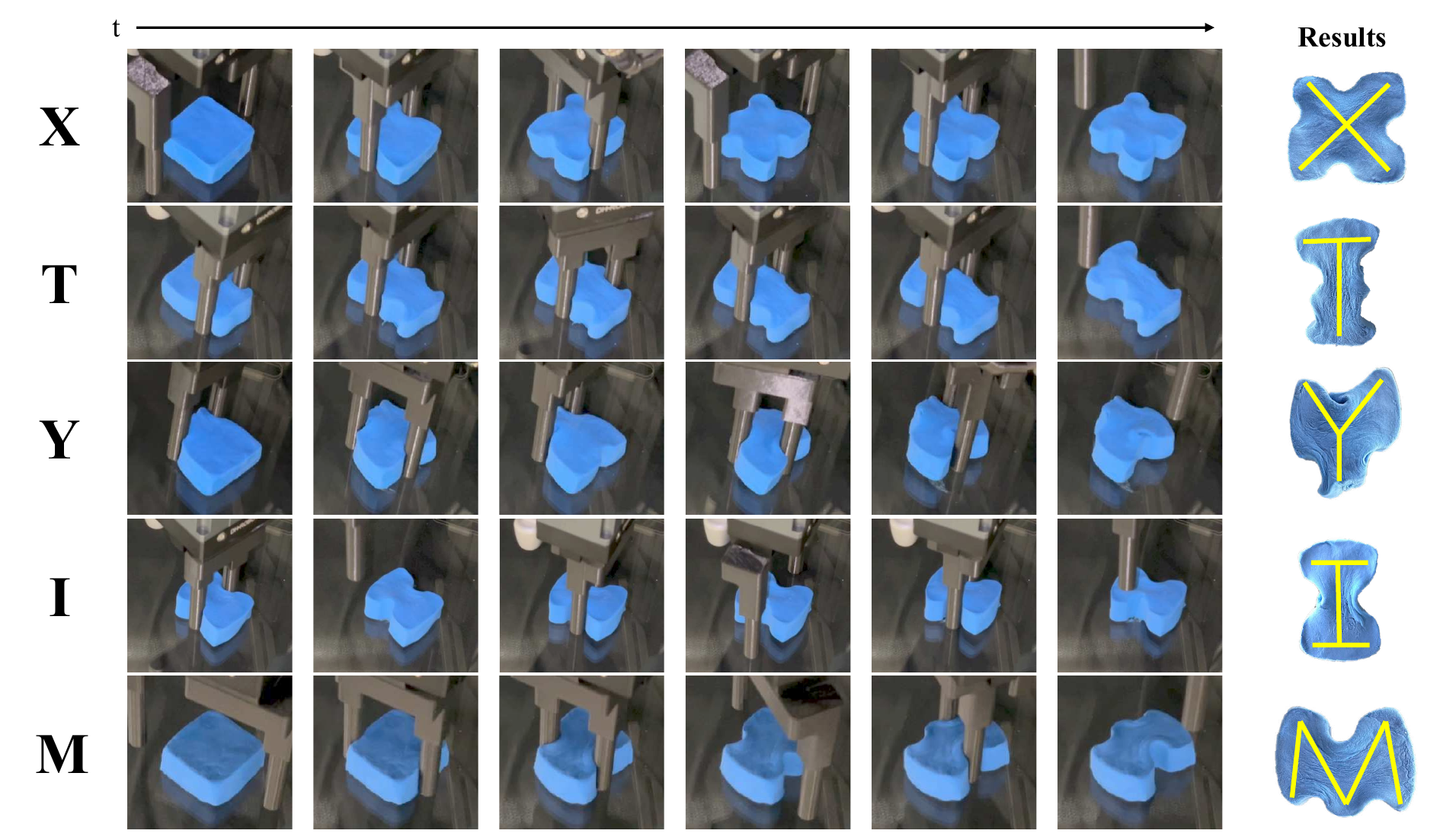}
    \caption{\textbf{Manipulation results of single-layer objects in real world.}} 
    \label{manipulate_result}
\end{figure}

\begin{figure}[tb]
    \centering
    \includegraphics[width=0.85\linewidth]{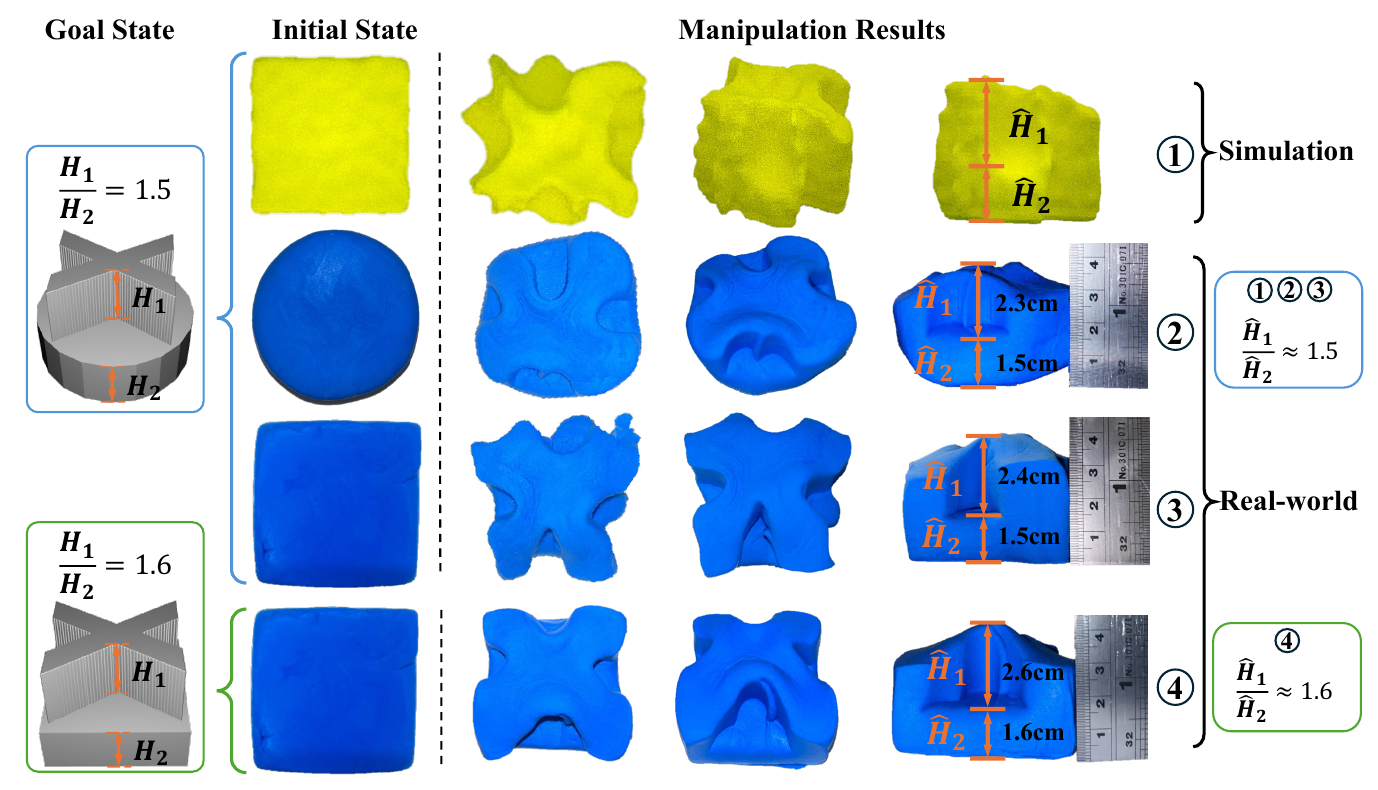}
    \caption{\textbf{Manipulation results of multi-layer objects in real world.} The initial heights of trials \ding{173}, \ding{174}, and \ding{175} are approximately $\SI{3.5}{\centi\meter}$, $\SI{3.0}{\centi\meter}$, and $\SI{3.2}{\centi\meter}$, respectively.
    }
    \vspace{-1ex}
    \label{3dcases}
\end{figure}

\begin{figure}[tb]
    \vspace{-3ex}
    \centering
    \includegraphics[width=0.9\linewidth]{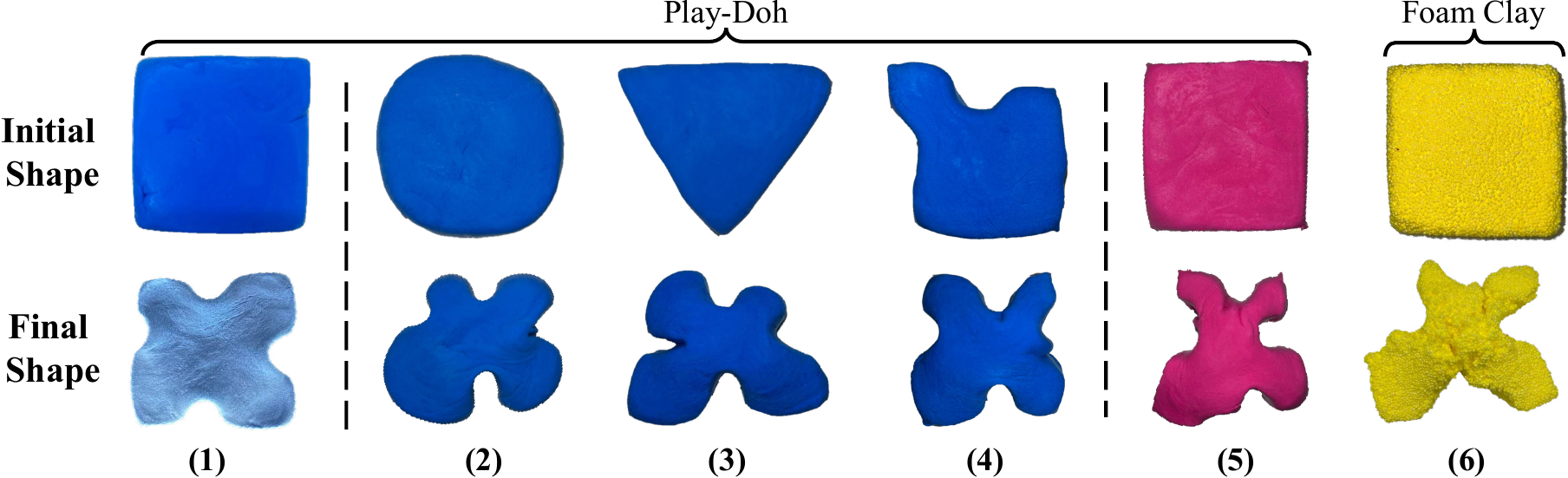}
    \caption{\textbf{Generalization demonstrations in terms of initial shapes, colors, and materials.} Six trials are conducted in four initial shapes, three colors, and two materials (dense play-doh vs. spongy form clay).}
    \label{diffinit}
\end{figure}

We conduct experiments both in the simulator and real world. We conduct three manipulation trials with letters ``X'', ``T'', and ``K'', and Fig.~\ref{sim_result} shows both qualitative and quantitative comparison results in the simulator with RoboCraft~\cite{robocraft} for fair comparison, showing the superior performance of our method. We present the manipulation results from real-world experiments in Fig.~\ref{manipulate_result}, showcasing five trials with varying levels of shaping difficulties utilizing a parallel two-finger gripper, such as the relatively easy case ``X'', which exhibits completely symmetrical (vertical, horizontal, and $45^{\circ}$ rotational symmetry), and the more challenging case ``Y'' and ``M'', which are symmetrical only about the vertical axis without rotational symmetry. Our proposed framework demonstrates the ability to successfully manipulate the plasticine into a shape closely resembling the goal shape. With our proposed framework, the robot can efficiently shape plasticine to meet different difficulty levels. To further justify the use of the 3D voxel-based representation, we design the experiments of pinching a two-layer stamp with a given height ratio $\frac{H1}{H2}$ between layers. As shown in Fig.~\ref{3dcases}, our proposed framework successfully pinches out the given two-layer stamps with a height ratio close to the desired. To show our method's generalization in terms of initial shapes, colors, and materials, we test the method in three more initial shapes (circle, triangle, and irregular polygon), two more colors (pink and yellow), and one more material (foam clay), as shown in Fig.~\ref{diffinit}. All final shapes are close to our goal, the ``X`` shape.

\begin{figure}[tb]
    \centering
    \includegraphics[width=0.8\linewidth]{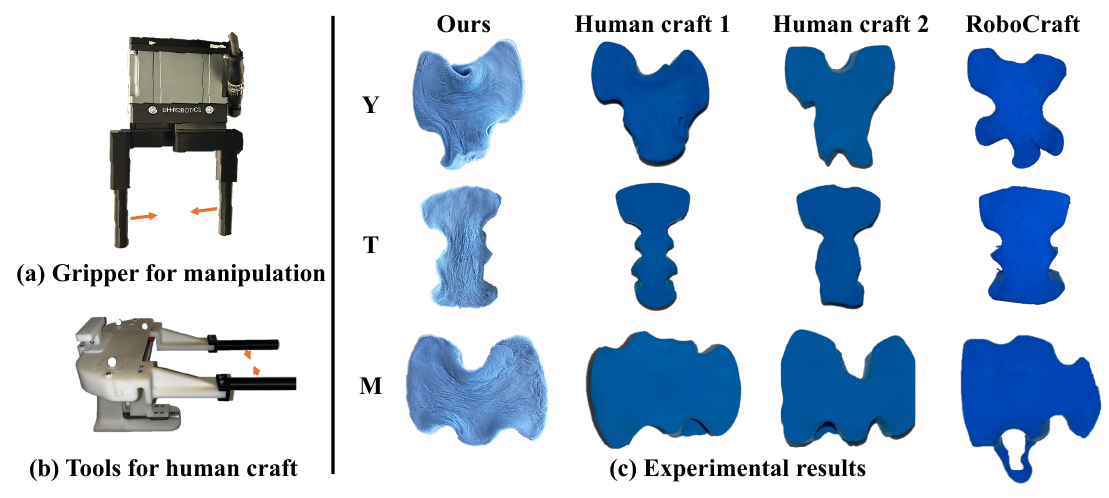}
    \caption{\textbf{Comparison of experimental results.} (a) The gripper for robot manipulation. (b) Tool for human manipulation inspired by~\cite{umi}. (c) The shaping results of our method, two amateur humans, and RobotCraft \cite{robocraft}.}
    \label{humandem}
\end{figure}

Furthermore, manipulation experiments of RoboCraft~\cite{robocraft} are conducted in the real world, and human manipulation trials are conducted using a 3D-printed device (Fig.\ref{humandem} (b)), which operates in the same manner as the gripper (Fig.\ref{humandem} (a)). Amateurs are allowed to pinch the plasticine the same number of times as the robot. As shown in Fig.\ref{humandem} (c), our method achieves comparable performance to human demonstrations and outperforms RoboCraft~\cite{robocraft}.

\subsubsection{Ablation Study} 
\begin{figure}[tb]
    \centering
\includegraphics[width=0.85\linewidth]{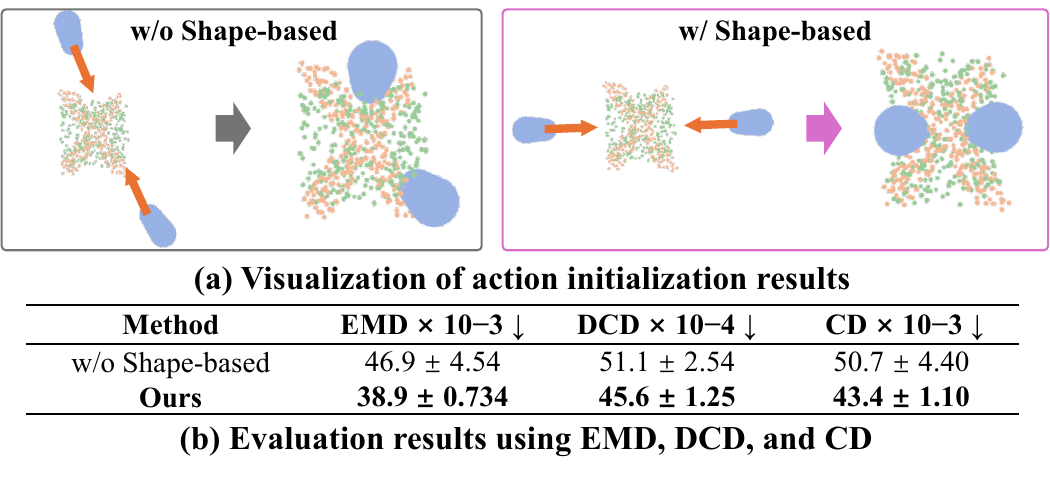}
    \caption{\textbf{Ablation study of shape-based action initialization.} “w/o shape-based” means that the actions are randomly initialized during manipulation. (a) Visualization of actions initialization results. (b) The evaluation results using EMD, DCD, and CD after the first action.}
    \label{shaperesult}
\end{figure}

To discern the contribution of proposed methods to the framework, we assess the contribution of the 3D CNN in modeling the dynamics,  contribution of shape-based action initialization in MPC, and the DCD loss.\\
\textbf{\textit{3D CNN Module:}}
Table~\ref{ablations} presents the ablation results for the 3D CNN module in the learned dynamics model. Without this module, the dynamics model is trained directly with the encoded state features of the sparse voxels. However, this straightforward approach performs worse than the method that integrates features derived from the 3D CNN module. Moreover, the ablation study indicates that the 3D CNN-based dynamics model achieves prior performance since it can effectively preserve multi-scale 3D spatial information.
\begin{table}[tb]
    \centering
    \renewcommand\arraystretch{1.0}
    \scalebox{0.9}{
    \begin{tabular}{cccc}
        \toprule
        \textbf{Method} & $\mathbf{EMD}_{\times10^{-3}}$~$\downarrow$ & $\mathbf{DCD}_{\times10^{-4}}$~$\downarrow$ & $\mathbf{CD}_{\times10^{-3}}$~$\downarrow$ \\
        \midrule
        w/o 3D CNN  & 24.5$\pm$5.07 & 27.2$\pm$2.25 & 34.3$\pm$2.60 \\
        w/o DCD loss & 23.4$\pm$3.94 & 26.7$\pm$3.96 & 34.2$\pm$2.36 \\
        \textbf{Ours} & $\mathbf{22.8\pm2.30}$ & $\mathbf{25.3\pm1.52}$ & $\mathbf{33.5\pm0.951}$\\
        \bottomrule
    \end{tabular}
    }
    \captionsetup{font=small} 
    \caption{\textbf{Ablation results in terms of the 3D CNN module and the DCD loss.} 
    }
    \label{ablations}
\end{table}
\\
\textbf{\textit{Shape-based Actions Initialization:}}
The results in Fig.~\ref{shaperesult} demonstrate the importance of using shape-based initialization. As shown in Fig.~\ref{shaperesult} (a), for the goal shape ``X'', without this module, the actions are randomly initialized, resulting in unreasonable and irreversible deformation. Additionally, we conduct experiments 10 times in identical experimental configurations with and without this module. We evaluate the shape similarity to the goal after the first action applied using EMD, DCD, and CD, and the quantitative results are shown in Fig.~\ref{shaperesult} (b). 
\\
\textbf{\textit{DCD Loss:}}
We conduct experiments of dynamics model training in identical experimental configurations with and without DCD loss. We evaluate the performance of the learned dynamics model, the results are shown in Table.~\ref{ablations}.
\section{Conclusion and Discussions}
In this paper, we propose a framework to manipulate the elasto-plastic object (e.g., plasticine) into a desired goal shape with 3D occupancy and a learning-based predictive control algorithm enhanced by a shape-based action initialization module. Meanwhile, we build a novel platform for full spatial data collection during manipulation and propose a pipeline to generate dense 3D occupancy ground truth and train a 3D occupancy prediction network supervised by the generated occupancy ground truth to infer the 3D occupancy during manipulation. Through our comprehensive experiments, we demonstrate the superiority of 3D occupancy-based state representation and the proposed framework can successfully manipulate the plasticine into a given 3D shape with different levels of shaping difficulties and multi-layer structures, and achieve comparable performance to humans.

Our current framework relies on full-space observations enabled by transparent plane; without it, occlusions significantly degrade the quality of mesh reconstruction for 3D occupancy generation. Given the limited quality of point clouds from employed RGB-D cameras, Poisson surface reconstruction struggles to capture fine topological changes such as splits and merges, making them difficult to model accurately. Compared to segmented observations that commonly require separate processing, jointly leveraging RGB and category features allows for more efficient and semantically meaningful feature extraction. Due to disparities in material properties between simulation and the real world, we find fine-tuning on real data is essential. Finally, the computational demands of our method require GPU acceleration, limiting deployment on CPU-only systems and resource-constrained applications.


\end{document}